\documentclass{applemlr}
\usepackage{amsmath}
\usepackage{enumerate} 
\usepackage{algorithm}
\usepackage{algpseudocode}
\usepackage{amsfonts}
\usepackage{amsthm}
\usepackage{newtxtt}
\usepackage{cleveref}
\usepackage{diagbox} 
\usepackage{colortbl}
\usepackage{amssymb}
\usepackage{xspace}
\usepackage{wrapfig}
\usepackage{adjustbox}
\usepackage{tabularx}
\usepackage{booktabs}
\usepackage{mathtools}
\usepackage{tikz}
\usepackage{enumitem}
\usepackage{silence}
\usepackage{dsfont}
\usepackage[table]{xcolor}
\usepackage[dvipsnames]{xcolor}
\usepackage{multirow}
\usepackage{makecell}
\usepackage{xfakebold}
\usepackage{tabularx}

\usepackage{colortbl}
\newlength\savewidth\newcommand\shline{\noalign{\global\savewidth\arrayrulewidth
  \global\arrayrulewidth 1pt}\hline\noalign{\global\arrayrulewidth\savewidth}}
\definecolor{baselinecolor}{gray}{.9}
\newcommand{\baseline}[1]{\cellcolor{baselinecolor}{#1}}

\newcommand{\method}{UniReasoner}

\usepackage{eccvabbrv}

\usepackage{amsmath,amsfonts,bm}

\def\eqref#1{equation~\ref{#1}}

\def\1{\bm{1}}

\DeclareMathAlphabet{\mathsfit}{\encodingdefault}{\sfdefault}{m}{sl}
\SetMathAlphabet{\mathsfit}{bold}{\encodingdefault}{\sfdefault}{bx}{n}

\definecolor{textgray}{HTML}{6E6E73}
\usetikzlibrary{positioning, calc}
\usetikzlibrary{decorations.pathmorphing}

\makeatletter
\patchcmd{\wrong@fontshape}{\@gobbletwo}{}{}{}
\makeatother
\makeatletter
\AtBeginDocument{%
  \urlstyle{sf}
  
}
\makeatother

\numberwithin{equation}{section} 
\tcbuselibrary{minted}
\setminted[python]{
  linenos,
  breaklines,
  fontsize=\footnotesize,
  xleftmargin=2em
}

\definecolor{light}{RGB}{125, 125, 125}
\crefname{tcb@cnt@pbox}{code}{code}
\Crefname{tcb@cnt@pbox}{Code}{Code}
\crefname{assumption}{assumption}{assumption}
\Crefname{assumption}{Assumption}{Assumptions}

\newtcolorbox[auto counter]{pbox}[2][]{
  colback=white,
  title=Code~\thetcbcounter: #2,
  #1,fonttitle=\sffamily,
  fontupper=\sffamily,
  arc=2pt,
  colframe=bgcolor,
  coltitle=fgcolor,
  colbacktitle=bgcolor,
  toptitle=0.25cm,
  bottomtitle=0.125cm
}

\makeatletter
\newcommand\applefootnote[1]{%
  \begingroup
  \renewcommand\thefootnote{}%
  \renewcommand\@makefntext[1]{\noindent##1}%
  \footnote{#1}%
  \addtocounter{footnote}{-1}%
  \endgroup
}
\makeatother

\definecolor{cverbbg}{gray}{0.90}

\title{Large Language Models are Universal Reasoners for Visual Generation}

\author{
\parbox{\textwidth}{
Sucheng Ren$^{1,2}$, Chen Chen$^2$, Zhenbang Wang$^2$, Liangchen Song$^2$, Xiangxin Zhu$^{2\star}$, Alan Yuille$^{1\star}$, Liang-Chieh Chen$^{2\star}$, Jiasen Lu$^{2\star}$ }}

\affiliation{$^1$Johns Hopkins University,~~ $^2$Apple}

\contribution{$^\star$Senior authors}

\abstract{

Text-to-image generation has advanced rapidly with diffusion models, progressing from CLIP and T5 conditioning to unified systems where a single LLM backbone handles both visual understanding and generation. 
Despite the architectural unification, these systems frequently fail to faithfully align complex prompts during synthesis, even though they remain highly accurate at verifying whether an image satisfies those same prompts.
We formalize this as the \emph{understanding-generation gap} and propose \method{}, a framework that leverages the LLM as a universal reasoner to convert its understanding strength into direct generation guidance.
Given a prompt, the LLM first produces a coarse visual draft composed of discrete vision tokens. It then performs a self-critique by evaluating the draft for prompt consistency, producing a grounded textual evaluation that pinpoints what needs to be corrected.
Finally, a diffusion model is conditioned jointly on the prompt, the visual draft, and the evaluation, ensuring that generation is guided by explicit corrective signals.
Each signal addresses a limitation of the other: the draft provides a concrete, scene-level anchor that reduces under-specification in text-only conditioning, while the evaluation turns verification into grounded, actionable constraints that correct omissions, hallucinations, and relational errors. Experiments show that \method{} improves compositional alignment and semantic faithfulness under the same diffusion backbone while maintaining image quality, demonstrating a practical way to exploit LLM reasoning to close the understanding–generation gap.

}

\date{\sffamily\today}

\begin{document}

\maketitle

\section{Introduction}

Text-to-image generation~\citep{sd,sd3,flux,dalle3,sana,emu3} has rapidly advanced with diffusion models, enabling photorealistic image generation at scale.
Early large-scale systems (\eg, Stable Diffusion~\citep{sd}) typically condition a latent diffusion backbone on a frozen text encoder, most prominently CLIP~\citep{clip} text embeddings, effectively treating language as a static conditioning signal. 
Subsequent diffusion Transformers~\citep{dit}, like SD3~\citep{sd3} and FLUX~\citep{flux}, strengthen the text conditioning by incorporating higher-capacity language encoders such as T5~\citep{t5}.
Yet, the fundamental paradigm remains largely unchanged: the prompt is compressed into a single dense embedding, and the diffusion model is tasked with satisfying all semantic and compositional constraints derived solely from that representation.

Recently, the field is increasingly shifting toward LLM-conditioned image generation~\citep{gpt4o,dalle3}. OpenAI’s GPT-4o~\citep{gpt4o} enables image generation “by chatting” with GPT-4o itself, reflecting a broader move toward multimodal LLMs as the primary interface for visual creation. 
Driven by the rapid progress of large language models, a new unification of understanding and generation systems~\citep{qwen,qwenimage,bagel,blip3o,chen2025janus,emu3, tian2025unigen, tian2025unigen1.5, lu2023uio2} has emerged, where a single LLM backbone supports both visual understanding and visual generation. 
Representative frameworks, such as BAGEL~\citep{bagel}, bridge understanding and generation with the same underlying LLM (\eg, Qwen~\citep{qwen}).
While this architectural unification is a significant step forward that injects deep semantic reasoning into the generative process, it does not fully resolve prompt-image inconsistencies. Even when the resulting images exhibit high perceptual quality, they frequently fail to faithfully satisfy complex, multi-constraint specifications.

\begin{figure}[t]
    \centering
    \includegraphics[width=\linewidth]{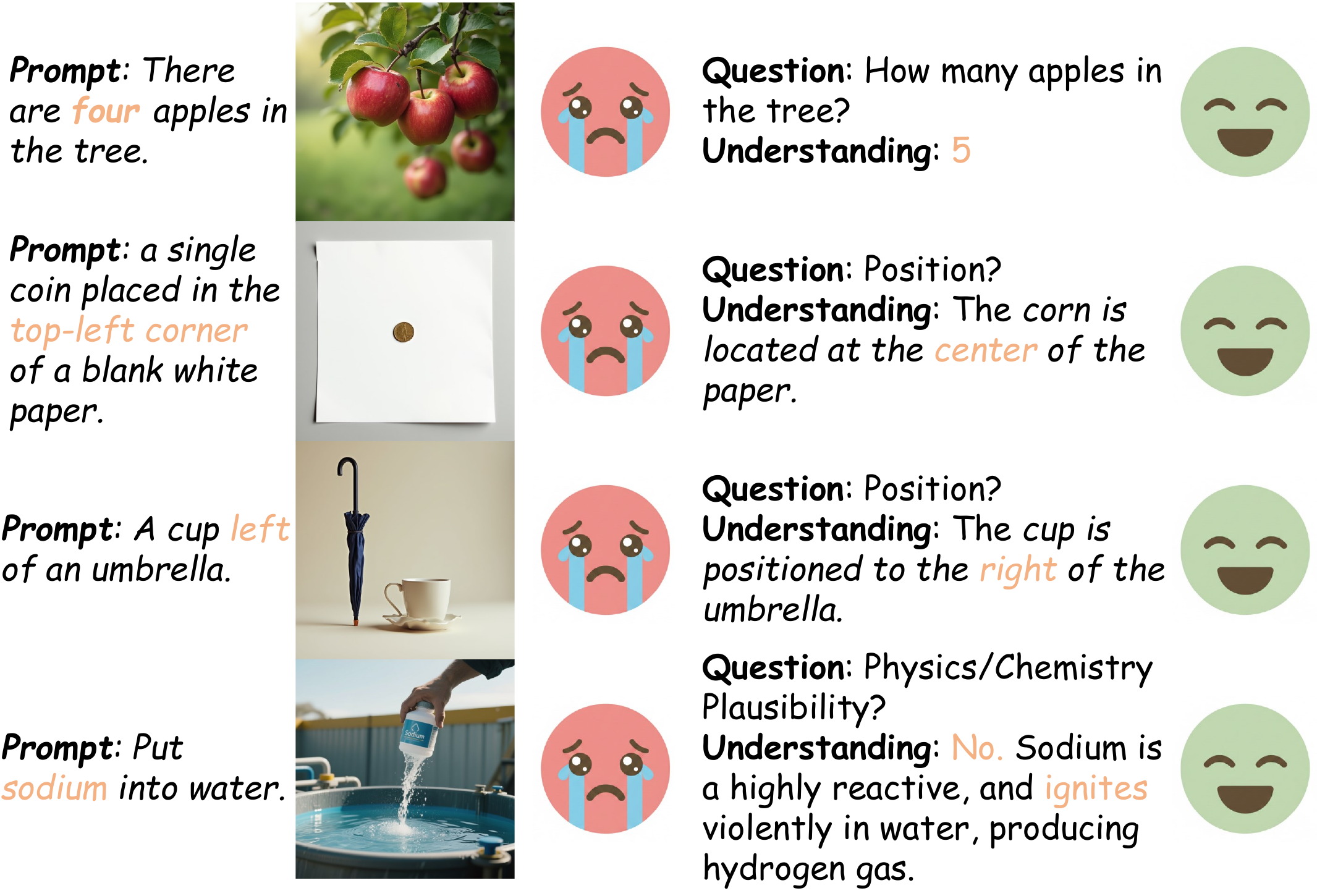}
\caption{\textbf{The Understanding-Generation Gap.} BAGEL~\citep{bagel}, employing the same LLM (Qwen~\citep{qwen}) for image generation and understanding, exposes a striking asymmetry. During generation, the model violates explicit prompt constraints, resulting in incorrect object counts, swapped spatial relations, or physically/chemically implausible outcomes.
However, when tasked with evaluating its own output, the exact same model accurately diagnoses these failures, demonstrating that its understanding strength exceeds its direct generative capabilities.}
    \label{fig:observation}
\end{figure}

A key observation motivating our work is that unified models~\citep{bagel,blip3o} with the same LLM for both understanding and generation exhibit a \textit{understanding–generation gap}. When asked to generate images that satisfy complex prompts, these models often produce plausible-looking outputs that nevertheless deviate from the specification. However, when tasked with verifying whether a given image matches that same prompt, they are substantially more dependable. As shown in Figure~\ref{fig:observation}, using BAGEL~\citep{bagel} for both generation and understanding exposes a striking asymmetry: the model generates five apples when prompted for four, yet correctly counts the resulting apples when tasked to evaluate the image.
Similar failures appear consistently across spatial relations and physical plausibility. These failure modes are often easy for the same model to diagnose after the fact -- suggesting that evaluation is a stronger primitive than direct generation, and that we should explicitly convert this verification strength into actionable guidance for diffusion synthesis. 

Motivated by this insight, we propose \method{}, a framework that leverages the LLM as a universal reasoner, converting its evaluation ability and internal knowledge into explicit generation control via a \emph{Draft-Evaluate-Diffuse} pipeline.
Given a prompt, the LLM first produces a \emph{visual draft} composed of discrete vision tokens, serving as a coarse but scene-level plan of the intended output.
Crucially, we do not treat this draft as the final image.
Instead, the same LLM evaluates the draft against the original prompt to produce a \emph{grounded evaluation}, a concise textual description of what needs to be corrected.
We then condition a diffusion model jointly on the original prompt, the visual draft, and the textual evaluation.
Consequently, the generation is guided by explicit corrective signals rather than relying on a single pass to implicitly capture all constraints.

Importantly, compared to the standard text-encoding and diffusion pipeline~\citep{sd,sd3}, our approach provides a more informative conditioning and each signal addresses the limitations of the other: blindly following the draft would preserve its mistakes, while evaluating without a visual reference reduces the reasoning process to transitional prompt rewriting.
By pairing them, the visual draft provides the evaluator with a concrete spatial anchor to critique, and the evaluation supplies the generator with localized, actionable instructions on ``what-to-fix.'' Together, this synergy turns the LLM's stronger understanding ability into direct generation guidance, improving compositional alignment without requiring structural changes to the underlying diffusion backbone. Extensive experiments on Text-to-Image (T2I) synthesis validate our approach: utilizing the same frozen SANA~\citep{sana} diffusion model, \method{} improves overall performance from 0.79 to 0.88 on GenEval~\citep{sana}, and from 84.50 to 86.30 on DPG-Bench~\citep{dpg}.

\section{Related Work}

\subsection{Text-to-Image Generation}
Diffusion models~\citep{ddpm,ddim,elucidating,score,glide,imagen,parti,sd,sd3,flux,dalle3,gpt4o,dit,sit} have become the dominant paradigm for text-to-image generation.
Stable Diffusion~\citep{sd} popularized latent diffusion with CLIP~\citep{clip} conditioning, and subsequent systems like FLUX~\citep{flux} and SD3~\citep{sd3} adopt Transformer-based denoisers~\citep{dit} with higher-capacity encoders such as T5~\citep{t5}.
A more recent trend involves augmenting or replacing classic text encoders entirely with Large Language Model (LLM) backbones. Qwen-Image~\citep{qwenimage}, for example, integrates the Qwen~\citep{qwen} LLM as a highly capable conditioning backbone for diffusion generation. Concurrently, the field is shifting toward unified multimodal foundation models -- such as BAGEL~\citep{bagel}, BLIP3-o~\citep{blip3o}, and Janus-Pro~\citep{chen2025janus} -- that aim to support both visual understanding and generation within a single system, often by blending autoregressive token modeling with diffusion or flow-matching components. 
Despite this progress, prompt-image inconsistencies remain prevalent for fine-grained constraints like counting, spatial relations, and attribute binding, suggesting that stronger backbones alone do not close the gap between understanding a specification and reliably generating an image that satisfies it. In contrast, our \method{} reform the LLM as a universal reasoner instead of only architecture-level unification. We convert its inherent evaluation strength into an explicit generation-time signal, using a prompt, a visual draft and a grounded evaluation to guide diffusion toward targeted corrections rather than relying on a single text embedding.

\subsection{Reasoning and Refinement for Visual Generation}

LLM reasoning has been applied to generation primarily as a front-end that reformulates prompts into generator-friendly conditions.
Common approaches include prompt rewriting~\citep{dalle3,chatgpt}, recaptioning with chain-of-thought planning~\citep{yang2024mastering}, and spatial layout generation via bounding boxes or scene blueprints~\citep{layoutgpt,gani2024llm,lian2023llm,he2025plangen}.
These methods reason entirely in text or coordinate space -- none produce a visual representation of the scene or verify the plan before generation.

A parallel line of work improves generation through
post-hoc verification or refinement at inference time.
UniGen~\citep{tian2025unigen} uses the same model as both
generator and verifier, applying text-based Chain-of-Thought
Verification for Best-of-N selection.
SLD~\citep{wu2024self} uses an LLM and a detector to diagnose
mismatches, then performs latent-space edits.
Reflect-DiT~\citep{li2025reflect} has a VLM critique for each
image and conditions subsequent generations on past feedback.
All these methods reason exclusively in text or pixel space.
None produces intermediate visual tokens as a draft
representation that can be both evaluated and fed as
conditioning to the generator in a single pass.

Our \method{} differs along three axes: (i) we reason in a multimodal token space -- the LLM produces a visual draft as discrete tokens rather than text or coordinates; (ii) we provide corrective guidance \emph{before} generation via single-pass conditioning, avoiding iterative regeneration; and (iii) the same LLM serves as both drafter and evaluator, keeping the pipeline self-contained.

\section{Method}
\label{sec:method}
\begin{figure}[t]
    \centering
    \includegraphics[width=\linewidth]{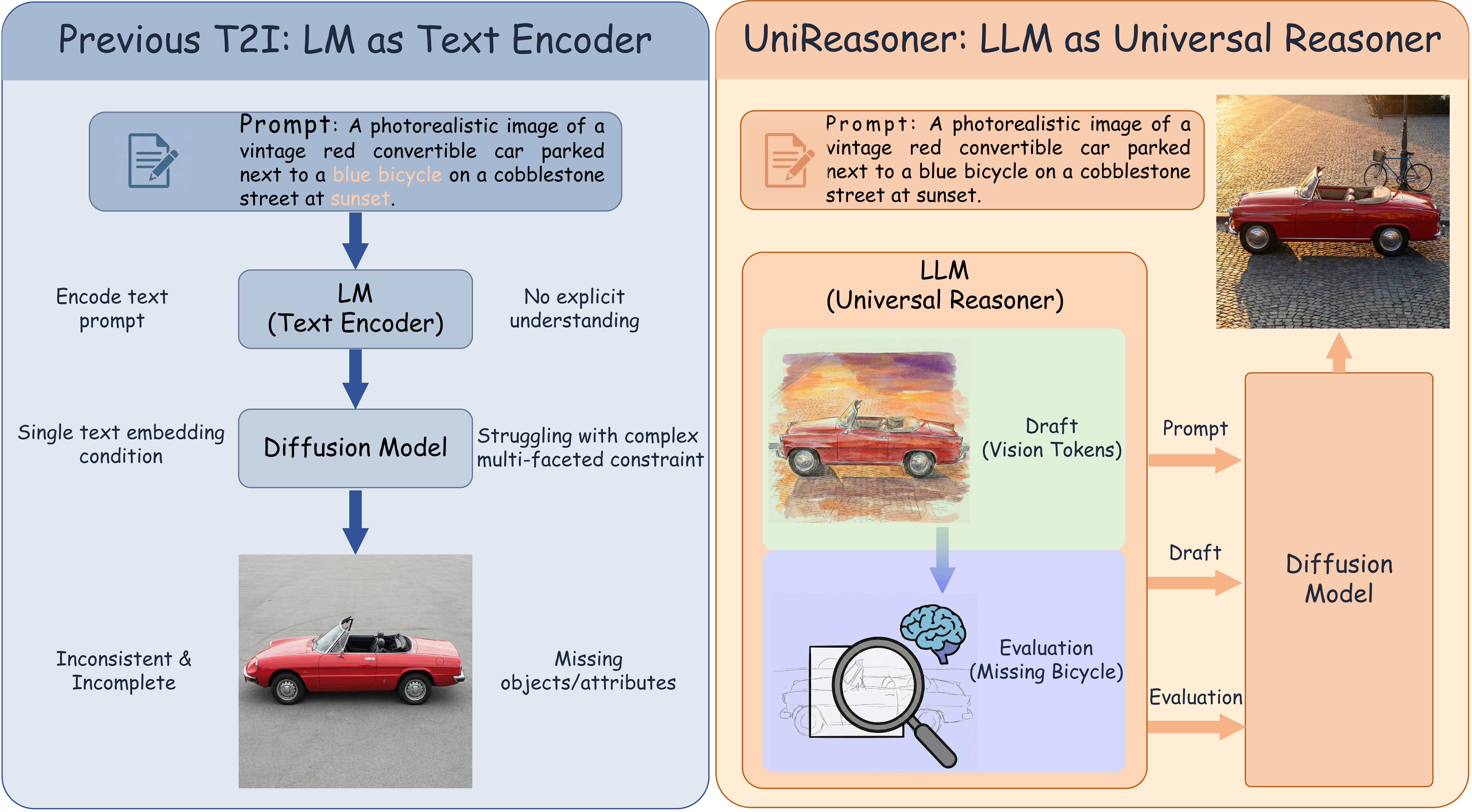}
    \caption{\textbf{Overview of \method{}.}  \textbf{Left:} Prior text-to-image pipelines utilize a Language Model (LM) (\eg, T5) solely as a text encoder, conditioning the diffusion model on a single embedding of the text prompt $p$. This often fails to satisfy complex prompts, leading to omissions or relational errors.
    \textbf{Right:} The proposed \method{} treats an LLM as a universal reasoner via a \textit{Draft--Evaluate--Diffuse} pipeline.
    It first generates a discrete visual draft $d$ to establish a spatial plan, then performs a self-critique to produce a grounded evaluation $e$ identifying prompt-draft discrepancies. Finally, the diffusion model is conditioned on the joint triplet $(p, d, e)$, transforming LLM's verification strength into explicit corrective signals during synthesis.
    }
    \label{fig:method}
\end{figure}

\textbf{Overview.} We study the task of text-to-image synthesis: given a text prompt $p$, the goal is to generate an image $I$ that is both perceptually high-quality and semantically faithful to $p$.
Following the recent trend of utilizing a single LLM for both understanding and generation, we propose \method{} to alleviate the observed \textit{understanding-generation gap} via a three-stage \textit{Draft-Evaluate-Diffuse} reasoning pipeline:

\begin{equation}
\label{eq:ded_pipeline}
d \sim \mathrm{Draft}_{\phi}(p), \qquad
e = \mathrm{Eval}_{\phi}(p, d), \qquad
I \sim \mathrm{Diffuse}_{\theta}(p, d, e),
\end{equation}
where $d$ is a visual draft represented as discrete vision tokens (serving as a coarse visual plan), $e$ is a grounded evaluation describing discrepancies between the prompt $p$ and the draft $d$, and $\mathrm{Diffuse}_{\theta}$ is a diffusion model (parameterized by $\theta$) conditioned on the joint tuple $(p, d, e)$, enabling targeted corrections during denoising by leveraging the explicit feedback provided in the draft and evaluation. Notably, both the drafting and evaluation stages are executed by the same underlying LLM (parameterized by $\phi$), framing it as a universal reasoner for visual synthesis. We detail each stage of this pipeline below.
 
\subsection{Visual Drafting via LLM Token Generation}
\label{sec:method_draft}

The first stage of our pipeline constructs a \textit{visual draft} $d$ that serves as a coarse spatial and semantic plan. We derive our draft space from SigLIP 2~\citep{siglip} features, which are optimized for semantic understanding and prompt-image alignment.

Let $F(I^d) \in \mathbb{R}^{H \times W \times C}$ denote the feature map extracted from a reference draft image $I^d$ (data preparation is detailed in Section~\ref{sec:data}). While continuous features provide rich information, they are difficult to sample and incompatible with the autoregressive generation typical of LLMs.

To resolve this, we discretize the SigLIP 2 features using Vector Quantization (VQ)~\citep{vq, tatok}. By mapping continuous features to a codebook of $K$ discrete indices, we obtain a representation that is both sampling-friendly and natively compatible with the LLM's vocabulary.

\textbf{Why SigLIP-based Discretization?} Unlike traditional pixel-reconstruction codebooks (\eg, VQGAN~\citep{esser2021taming}), SigLIP-quantized tokens encode high-level semantic primitives. This ensures the draft space is inherently aligned with the LLM’s internal world knowledge, making the tokens more ``readable'' for the subsequent self-critique stage.

\textbf{Drafting via Token Generation.}
Given a prompt $p$, the LLM (parameterized by $\phi$) generates the visual draft $d$ as a sequence of discrete tokens:
\begin{equation}
\label{eq:draft_ar}
d \sim p_{\phi}(d \mid p).
\end{equation}Concretely, we represent each VQ index $k \in \{1, \dots, K\}$ as a unique special token $\langle v_k \rangle$ within the LLM's expanded vocabulary. The draft is generated as a contiguous block within a task-specific wrapper:
\begin{equation}
\label{eq:draft_format}\langle \mathrm{DRAFT} \rangle \ \langle v_{q_1} \rangle \cdots \langle v_{q_N} \rangle \ \langle /\mathrm{DRAFT} \rangle,
\end{equation}
where $N$ is the number of tokens in the spatial grid. This allows the LLM to ``visualize'' the scene within its native generative interface. Intuitively, this process converts an underspecified linguistic description into a concrete visual anchor, reducing the ambiguity that often plagues single-pass prompt embeddings. The LLM is trained to generate these tokens using a standard cross-entropy objective:
\begin{equation}
\label{eq:loss_draft}
\mathcal{L}_{\text{draft}}=-\sum_{i=1}^{N}\log p_{\phi}(q_i\mid p,q_{<i}).
\end{equation}

\subsection{Grounded Evaluation via LLM Self-Critique}
\label{sec:method_eval}

While visual drafting provides a concrete spatial anchor, it does not inherently guarantee prompt faithfulness. The pivotal step in our pipeline is the conversion of the LLM’s internal verification strength into explicit, actionable guidance. We task the same LLM (parameterized by $\phi$) with evaluating the draft $d$ against the original prompt $p$:
\begin{equation}
\label{eq:eval}
e = \mathrm{Eval}_{\phi}(p,d).
\end{equation}

\textbf{Grounding via Self-Critique.}
To perform this self-critique, the LLM is provided with (i) the original prompt $p$, (ii) the discrete visual draft $d$, and (iii) instructions to identify semantic inconsistencies or violations of visual commonsense. The resulting output $e$ is a \textit{grounded evaluation} that explicitly pinpoints specific mismatches rather than providing a generic caption or a simple prompt rewrite.
This grounding is critical: conditioning a generator solely on the pair $(p,d)$ would inadvertently encourage the model to preserve errors present in the draft. By contrast, the grounded evaluation $e$ instructs the generator on exactly where and how the draft deviates from the prompt. This allows the downstream diffusion model to treat the draft $d$ as a proposed spatial layout and the evaluation $e$ as a set of corrective constraints, transforming the LLM's understanding strength into a diagnostic text stream that enables targeted semantic correction during synthesis.

\subsection{Image Synthesis via Joint Diffusion Conditioning}
\label{sec:method_diffuse}

The final stage of our pipeline generates the image $I$ by conditioning a diffusion model on the triplet $(p, d, e)$. Let $z_t$ denote the noisy latent at timestep $t$, and let $\epsilon_{\theta}$ be the noise predictor. While standard text-to-image models condition only on a prompt embedding $c(p)$:
\begin{equation}
\label{eq:diff_textonly}
\epsilon_{\theta}(z_t, t; c(p)),
\end{equation}
we instead construct a multi-source conditioning signal by concatenating the prompt, visual draft, and grounded evaluation:
\begin{equation}
\label{eq:cond_concat}
c(p, d, e) = c\big(\mathrm{Concat}(p, d, e)\big).
\end{equation}
The denoising process then proceeds as:
\begin{equation}
\label{eq:diff_multicond}\epsilon_{\theta}(z_t, t; c(p, d, e)).
\end{equation}
Here, $c(\cdot)$ represents the LLM used to encode the joint sequence into a unified feature space. These features are injected into the diffusion backbone via MM-DiT~\citep{sd3} or cross-attention layers~\citep{sana}, depending on the specific architecture.

\begin{figure*}[!t]
    \centering
    \includegraphics[width=\linewidth]{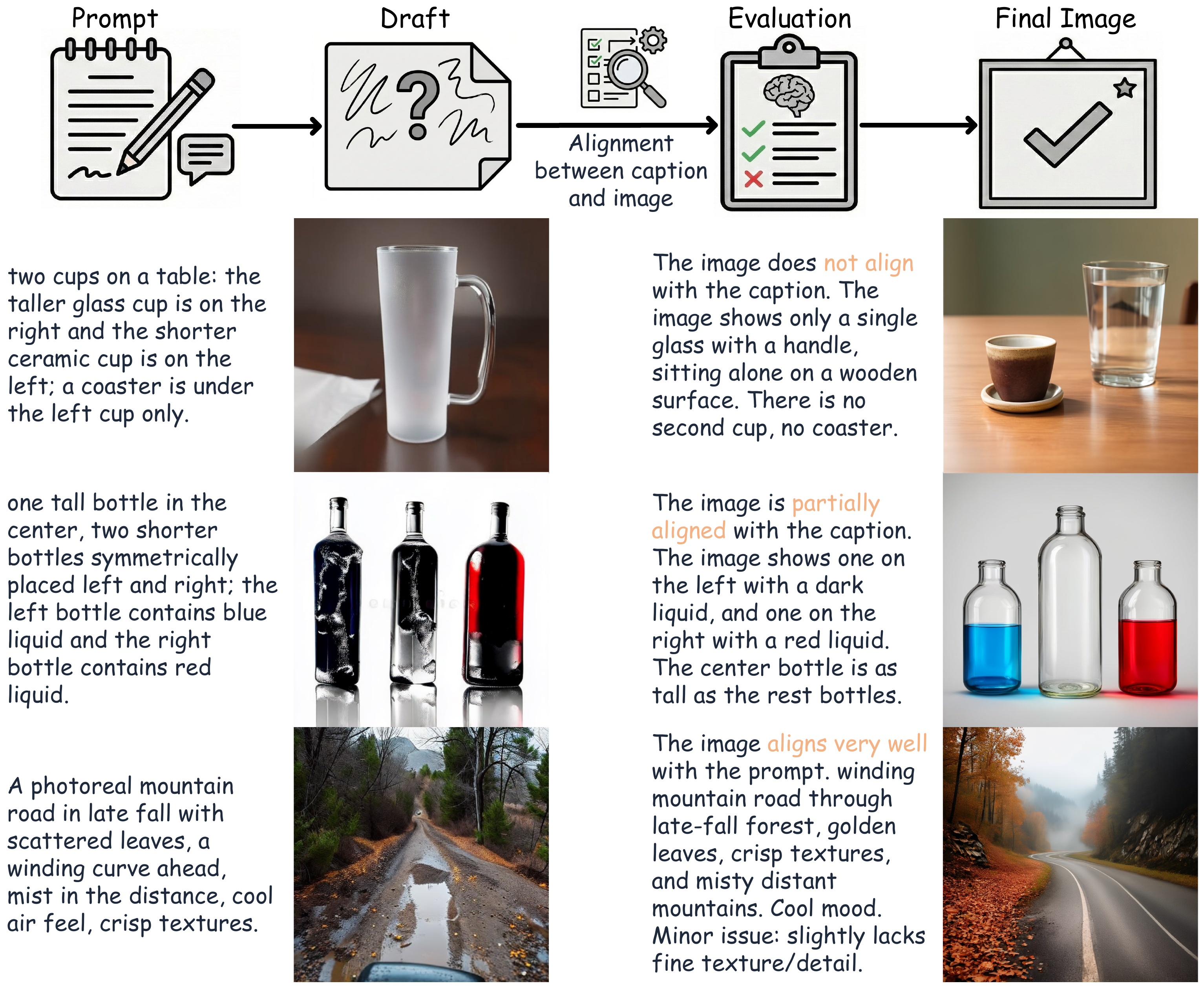}
\caption{\textbf{Qualitative Results of \method{}.}
Given a text prompt, it first generates a coarse visual draft that provides a semantically grounded plan.
It then evaluates prompt--draft alignment and explicitly describes unsatisfied constraints (\eg, missing objects, incorrect counts/attributes, or erroneous spatial relations). Finally, a diffusion model synthesizes the image conditioned jointly on \{prompt, draft, evaluation\}, using the evaluation as an explicit ``what-to-fix'' signal to correct draft errors and ensure compositional faithfulness. Note that while the draft exists only as discrete tokens during synthesis, we decode them here for visualization.
}
    \label{fig:evaluate}
\end{figure*}

\textbf{Joint Diffusion Conditioning.}
Compared to standard text-only pipelines, our multi-source conditioning provides two complementary signals: (i) the visual draft $d$ supplies a semantically grounded spatial plan, reducing linguistic ambiguity and preserving complex interacting constraints; and (ii) the grounded evaluation $e$ provides explicit corrective guidance on what must be resolved. By combining these, the diffusion model can allocate its generative capacity toward specific, localized mismatches rather than implicitly attempting to resolve all constraints from a single, potentially diluted prompt embedding. Together, drafting and evaluation transform LLM's verification strength into structured, generation-time control, yielding significantly more faithful results without requiring architectural changes to the diffusion backbone. 

\textbf{Visualization of the Reasoning Process.}
In Figure~\ref{fig:evaluate}, we illustrate the whole reasoning process of \method{}, including draft, evaluation and the final generated image. 
The visual draft and the grounded evaluation together transform the LLM's  understanding strength into an explicit diagnostic stream, pinpointing the precise semantic corrections required during diffusion refinement.

\section{Experimental Results}
\subsection{Experimental Setup}
\subsubsection{Dataset Construction.} 
\label{sec:data}
To train our \method{} framework, we construct a three-tuple training signal $(p, d, e)$ paired with a target image $I^t$. Here, $p$ is the original text prompt, $d$ represents the visual draft (a sequence of imperfect but semantically informative tokens capturing an initial  plan), and $e$ is a grounded evaluation produced by a Vision-Language Model (VLM)\footnote{The VLM is used exclusively for offline dataset construction; our framework \method{} relies solely on the base LLM as the universal reasoner.} that explicitly diagnoses the prompt-draft alignment. Our training protocol consists of two distinct stages: (i) large-scale pretraining using reconstructed images to establish baseline reasoning, and (ii) targeted finetuning using model-generated hard negatives to refine corrective capabilities. We detail each stage below.

\textbf{Stage I: Pretraining via Image Reconstruction.}
We utilize the same text-image dataset as prior works~\citep{blip3o,uniworld} containing only $(p, I)$ pairs. Because it lacks intermediate visual drafts, we synthesize them via token reconstruction:
\begin{itemize}
  \item \textbf{Draft Supervision.} For each image $I$, we obtain a degraded reconstruction $\tilde{I}$ using a pretrained image tokenizer~\citep{emu3}. We treat $\tilde{I}$ as the reference draft image $I^d$ and discretize it into draft tokens $d$ via our SigLIP-based tokenizer.
  \item \textbf{Target Supervision.} The original, high-fidelity image $I$ serves as the final target $I^t$ for the diffusion model.
  \item \textbf{Grounded Evaluation.} We process the pair $(p, \tilde{I})$ through a VLM (Qwen-VL~\citep{qwenvl}) to generate the evaluation $e$. This evaluation checks for semantic consistency and verbalizes concrete mismatches (\eg, missing objects, swapped attributes, or incorrect spatial relations), yielding an explicit ``what-to-fix'' diagnostic text for the diffusion model.
\end{itemize}

\textbf{Stage II: Finetuning via Hard-Negative Candidates.}
To strengthen the model's ability to correct structural errors, we construct a curated finetuning set~\citep{sharegpt4o,blip3o} containing challenging prompt-draft mismatches:
\begin{itemize}
  \item \textbf{Candidate Generation.} For a given prompt $p$, we generate a candidate image $I^f$ using a state-of-the-art diffusion model FLUX~\citep{flux}.
  \item \textbf{Alignment Scoring.} We use Qwen-VL to score semantic alignment between the prompt $p$ and both the generated candidate $I^f$ and the real image $I$.
  \item \textbf{Hard-Negative Mining.} We select the poorly aligned candidate as the draft image $I^d$ (converted to tokens $d$) and the strictly better-aligned image as the final target $I^t$.
  \item \textbf{Evaluations for Correction.} Similar to Stage I, we generate the grounded evaluation $e$ by prompting the VLM to diagnose the discrepancies between $p$ and $I^d$.
\end{itemize}

\subsubsection{Implementation Details.} 
We instantiate our LLM backbone with Qwen~\citep{qwen}\footnote{We strictly use LLM backbone (Qwen~\citep{qwen}) during \method{} training and inference to ensure a fair comparison with text-to-image baselines that rely solely on language models.} and utilize SANA~\citep{sana} as the diffusion generator. To isolate the contribution of the LLM as a universal reasoner, we freeze the diffusion backbone entirely, optimizing only the LLM and the cross-modal connector linking the language model to the generator. We train the network using the AdamW optimizer~\citep{adamw} with an initial learning rate of $5\times10^{-5}$, applying a 1,000-step linear warmup followed by a decay schedule down to $1\times10^{-5}$. The model is pretrained for 60,000 iterations on the reconstructed dataset (Stage I) and subsequently finetuned for 20,000 iterations on the hard-negative candidate set (Stage II). We evaluate \method{}'s compositional faithfulness on  GenEval~\citep{ghosh2023geneval} and DPG-Bench~\citep{dpg}.

\begin{table*}[t]
    \centering
    \caption{
    \textbf{Evaluation of Text-to-Image Generation on Geneval.} Note that our \method{} and the SANA baseline share the exact same diffusion generator.
    }
   \resizebox{\linewidth}{!}{
    \begin{tabular}{c|c|cccccc}
    Method&  Overall& Single Obj.& Two Obj.& Counting& Colors& Position& Attr. Binding \\
    \shline
    Emu3~\citep{emu3}&0.54& 0.98&0.71&0.34&0.81&0.17&0.21\\
    DALL$\cdot$E~3~\citep{dalle3} & 0.67&  0.96&  0.87&  0.47 & 0.83 & 0.43&  0.45 \\
    FLUX.1-Dev~\citep{flux} & 0.66 &0.98& 0.81 &0.74& 0.79& 0.22& 0.45 \\
    SD3~\citep{sd} & 0.71 & 0.98 & 0.89 & 0.73&  0.83&  0.34&  0.47 \\
    Janus-Pro~\citep{chen2025janus}&0.80&0.99&0.92&0.85&0.91&0.75&0.66 \\
    BLIP-3o\citep{blip3o}&0.83&0.99&0.92&0.74&0.86&0.77&0.67 \\
    GPT-4o~\citep{gpt4o} &0.84& 0.99& 0.92& 0.85& 0.92 &0.75& 0.61 \\
    \hline
    SANA~\citep{sana}&0.79& 0.98&0.93&0.78&0.88&0.62&0.57\\
    \method{}  & 0.88&0.99&0.94&0.90&0.92&0.83&0.72\\
    \end{tabular}}
    \label{tab:geneval}
\end{table*}

\subsection{Main Results}
\label{sec:main_results}
\textbf{GenEval.}
As shown in Table~\ref{tab:geneval}, \method{} achieves the best overall GenEval~\citep{ghosh2023geneval} score of 0.88, surpassing all evaluated baselines. Notably, this gain is obtained without altering the underlying diffusion generator: \method{} shares the exact same backbone as the SANA baseline~\citep{sana}, yet increases the overall score from 0.79 to 0.88 (+0.09). In particular, \method{} raises Counting from 0.78 to 0.90, Position from 0.62 to 0.83, and Attribute Binding from 0.57 to 0.72, while maintaining near-ceiling performance on Single-Object and Two-Object prompts (0.99 and 0.94, respectively). These trends indicate that employing the LLM as a universal reasoner provides highly effective visual drafts and corrective cues for complex entity interactions—areas often underspecified by text-only conditioning—thereby significantly improving prompt adherence.

Furthermore, the results reveal a consistent progression tied to increasingly capable language conditioning. LM-conditioned diffusion backbones, such as SD3~\citep{sd3} and FLUX.1-Dev~\citep{flux}, trail behind LLM-based generators like BLIP-3o~\citep{blip3o}. While the latter primarily benefit from stronger linguistic grounding, they do not explicitly perform text-side reasoning. GPT-4o~\citep{gpt4o} improves upon these models by pairing advanced language understanding with explicit text reasoning. Building on this progression, \method{} advances the state-of-the-art by fully utilizing the LLM as a universal reasoner that actively drafts and evaluates, rather than treating language conditioning as a static, one-shot prompt. Overall, these results suggest that integrating a universal reasoning framework is a complementary and practical enhancement for high-fidelity diffusion generators, yielding strictly better semantic alignment and constraint satisfaction while preserving visual quality.

\textbf{DPG-Bench.}
We further evaluate \method{} on the DPG-Bench~\citep{dpg}. As reported in Table~\ref{tab:dpg}, \method{} achieves an overall score of 86.30, outperforming previous methods including DALL$\cdot$E 3~\citep{dalle3}, SD3~\citep{sd3}, FLUX.1-Dev~\citep{flux}, Janus-Pro~\citep{chen2025janus}, Emu3~\citep{emu3}, and BLIP-3o~\citep{blip3o}. Crucially, \method{} improves upon SANA~\citep{sana} by +1.80 overall ($84.50\!\rightarrow\!86.30$) using the identical diffusion generator, demonstrating that the performance gain stems directly from our reasoning framework rather than a more powerful diffusion backbone.

Breaking this down by category, \method{} shows the most significant gains on Global instructions ($77.55\!\rightarrow\!92.46$), indicating that grounded evaluation provides effective, high-level corrective cues for holistic prompt intent and scene consistency. We also observe consistent improvements across fine-grained compositional aspects, including Entity ($89.85\!\rightarrow\!90.56$) and Attribute ($89.96\!\rightarrow\!91.11$), alongside highly competitive performance on Relation (90.65) and Other (89.84). Together, these results validate that \method{} generalizes well beyond GenEval benchmark, robustly enhancing text-to-image alignment across diverse instruction families.

\begin{table*}[t]
    \centering
    \caption{\textbf{Evaluation of Text-to-Image Generation on DPG-Bench.} Note that our \method{} and the SANA baseline share the exact same diffusion generator. }
    \begin{tabular}{c|c|ccccc}
    Method&  Overall&  Global&  Entity & Attribute&  Relation & Other \\
    \shline
    Emu3~\citep{emu3}&80.60 &85.21& 86.68& 86.84& 90.22& 83.15\\
    BLIP-3o\citep{blip3o}&82.27&88.63&89.11 &87.84&87.03&89.46\\
    DALL$\cdot$E~3~\citep{dalle3} & 83.50& 90.97& 89.61& 88.39& 90.58& 89.83\\
    FLUX.1-Dev~\citep{flux} &83.84& 74.35& 90.00& 88.96& 90.87& 88.33 \\
    SD3~\citep{sd} & 84.08 &87.90 &91.01& 88.83 &80.70& 88.68 \\
    Janus-Pro~\citep{chen2025janus}&84.19&86.90& 88.90& 89.40& 89.32& 89.48\\
\hline
    SANA~\citep{sana}&84.50& 77.55&89.85&89.96&89.19&91.74\\
    \method{}  & 86.30&92.46&90.56& 91.11&90.65&89.84\\
    \end{tabular}
    \label{tab:dpg}
\end{table*}

\begin{figure}[!t]
    \centering
    \includegraphics[width=\linewidth]{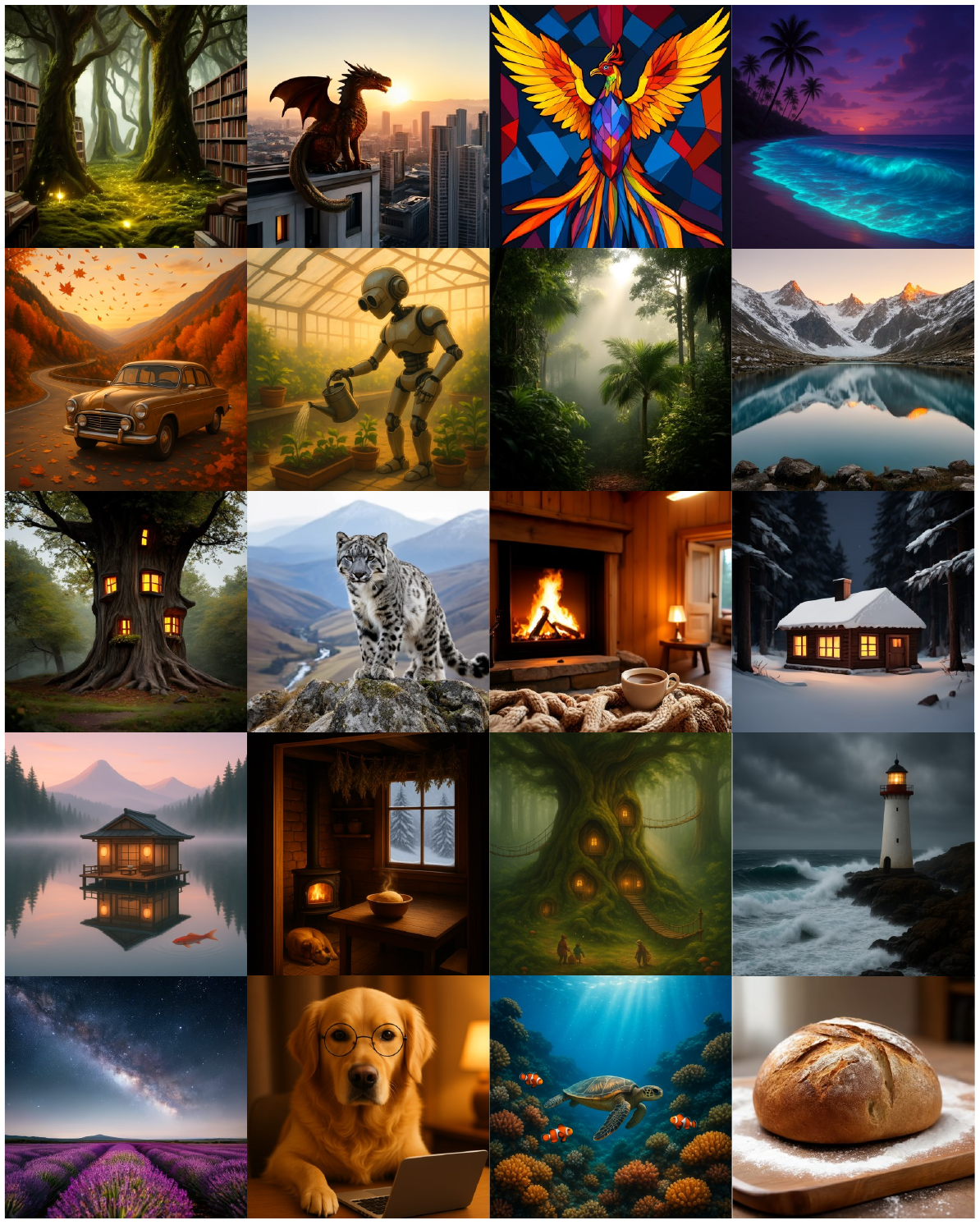}
    \caption{\textbf{Qualitative Results of \method{}.}
    We show images generated by \method{} across a diverse set of photorealistic and artistic prompts.
    }
    \label{fig:visualization}
\end{figure}

\textbf{Visualization.} We provide qualitative results of \method{} in Figure~\ref{fig:visualization}. 

\subsection{Ablation Study}
\textbf{Effectiveness of the LLM as a Universal Reasoner.}
Table~\ref{tab:reasoning} studies how the language backbone and the reasoning interface affect prompt faithfulness on GenEval~\citep{ghosh2023geneval}.
Replacing the standard T5 text encoder~\citep{t5} with a stronger LLM backbone (Qwen3~\citep{qwen}) already improves overall alignment from $0.70$ to $0.79$,
suggesting that better language understanding translates to better constraint adherence.
Adding text-only reasoning via prompt rewriting further brings consistent but modest gains for both backbones
(T5: $0.70\!\rightarrow\!0.76$, Qwen3: $0.79\!\rightarrow\!0.82$), indicating that rewriting reduces ambiguity yet remains limited when multiple constraints interact.

Transitioning to our full universal reasoning framework yields the largest improvement, boosting the Qwen3 text-only reasoning baseline from 0.82 to 0.88 overall, again without modifying the diffusion generator.
The gains concentrate on compositional categories that require explicit multi-constraint satisfaction:
Counting increases from $0.72$ to $0.90$ (+$0.18$),
Position from $0.72$ to $0.83$ (+$0.11$),
and Attribute Binding from $0.64$ to $0.72$ (+$0.08$),
while maintaining near-ceiling performance on Single-Object/ Two-Object prompts ($0.99/0.94$).
These results support our hypothesis that moving beyond text-only reasoning to a universal reasoning by the draft and evaluation
provides more actionable, localized correction cues, enabling the diffusion model to resolve specific mismatches rather than relying on a single (rewritten) prompt embedding.

\begin{table*}[t]
    \centering
    \caption{
    \textbf{Ablation of Language Models and Reasoning on GenEval.}
    ``Text'' reasoning refers to prompt rewriting, while ``Universal'' is our reasoning framework.
    }
   \resizebox{\linewidth}{!}{
    \begin{tabular}{c|c|c|cccccc}
    Text Encoder& Reasoning& Overall& Single Obj.& Two Obj.& Counting& Colors& Position& Attr. Binding \\
    \shline
    T5 &N/A & 0.70 & 0.98 & 0.86& 0.50&0.85&0.45&0.46\\
    T5 &Text & 0.76& 0.98&0.88 &  0.58& 0.86&0.52&0.50 \\
    Qwen3 & N/A & 0.79&0.98&0.90 & 0.65&0.88&0.69& 0.61\\
    Qwen3 & Text& 0.82& 0.99&0.91 & 0.72& 0.90&0.72&0.64\\
    \baseline{Qwen3} & \baseline{Universal}  & \baseline{0.88}&\baseline{0.99}&\baseline{0.94}&\baseline{0.90}&\baseline{0.92}&\baseline{0.83}&\baseline{0.72}\\
    \end{tabular}}
    \label{tab:reasoning}
\end{table*}

\begin{table*}[t]
    \centering
\caption{
\textbf{Ablation of Conditioning Signals on GenEval.} \textsc{Text}, \textsc{Draft}, and \textsc{Eval} denote the text prompt, visual draft, and grounded evaluation, respectively.
}
   \resizebox{\linewidth}{!}{
    \begin{tabular}{ccc|c|cccccc}
     \multicolumn{3}{c|}{Condition}
& \multirow{2}{*}{Overall}
& \multirow{2}{*}{Single Obj.}
& \multirow{2}{*}{Two Obj.}
& \multirow{2}{*}{Counting}
& \multirow{2}{*}{Colors}
& \multirow{2}{*}{Position}
& \multirow{2}{*}{Attr. Binding} \\
\textsc{Text} & \textsc{Draft} & \textsc{Eval} &&&&& \\
    \shline
    $\checkmark$& & & 0.79&0.98&0.90 & 0.65&0.88&0.69& 0.61\\
    &$\checkmark$ & & 0.82&0.99&0.92&0.71&0.84&0.76&0.67 \\
    $\checkmark$&$\checkmark$ && 0.82&0.99&0.92&0.72&0.82&0.77&0.68\\
    \baseline{$\checkmark$}&\baseline{$\checkmark$}&\baseline{$\checkmark$} & \baseline{0.88} & \baseline{0.99} & \baseline{0.94} & \baseline{0.90} & \baseline{0.92} & \baseline{0.83} & \baseline{0.72}\\
    \end{tabular}}
    \label{tab:condition}
\end{table*}

\textbf{Effectiveness of the \method{} Conditioning.}
Table~\ref{tab:condition} ablates the conditioning signals used by \method{}, where \textsc{Text}, \textsc{Draft}, and \textsc{Eval} denote the text prompt, visual draft, and grounded evaluation, respectively.
Using only the text prompt (\textsc{Text}, Row 1) yields an overall GenEval score of $0.79$.
Replacing the text entirely with the visual draft (\textsc{Draft}, Row 2) improves performance to $0.82$, with clear gains on compositional constraints such as Counting ($0.65\!\rightarrow\!0.71$), Position ($0.69\!\rightarrow\!0.76$), and Attribute Binding ($0.61\!\rightarrow\!0.67$), suggesting that the draft provides a more explicit, spatially grounded plan than a single text embedding.

Combining text and draft (\textsc{Text} + \textsc{Draft}, Row 3) offers limited additional benefit over the draft alone, with the overall score remaining at $0.82$. This implies that the visual draft largely subsumes the constraint-relevant information in the prompt, while naive fusion can introduce minor interference (\eg, Colors $0.84\!\rightarrow\!0.82$).
In contrast, augmenting the conditions with the grounded evaluation (\textsc{Text} + \textsc{Draft} + \textsc{Eval}, Row 4) produces a substantial jump to $0.88$ overall.
The improvement is dominated by categories that require multi-constraint correction: Counting increases from $0.72$ to $0.90$ (+$0.18$), Position from $0.77$ to $0.83$ (+$0.06$), and Attribute Binding from $0.68$ to $0.72$ (+$0.04$), while Single-Object and Two-Object performance stays near ceiling ($0.99/0.94$).
These results validate the role of evaluation as an explicit ``what-to-fix'' signal: given a draft that may violate constraints, the grounded evaluation enables the diffusion generator to correct localized errors rather than relying on implicit text-only conditioning.

\textbf{Comparison of Visual Draft Variants.}
Table~\ref{tab:vq} compares different visual draft representations under the same Draft-Evaluate-Diffuse pipeline.
The ``None'' baseline corresponds to standard text-only conditioning (no visual draft), achieving an overall GenEval score of $0.79$.
Introducing a draft based on continuous VAE latents~\citep{flux} significantly degrades performance ($0.72$ overall), causing pronounced drops in compositional categories (\eg, Position $0.69\!\rightarrow\!0.54$ and Attribute Binding $0.61\!\rightarrow\!0.52$).
We attribute this to the fact that continuous VAE latents are ill-suited for the discrete autoregressive sampling required during the drafting phase.
Utilizing VQ tokens~\citep{emu3} yields better results ($0.84$ overall) compared to text-only conditioning (``None'' baseline), since discrete codes provide a more structured intermediate representation for autoregressive draft generation.
However, VQ still underperforms our final design.
Our SigLIP-based Discretization draft achieves the best performance ($0.88$ overall) and consistently dominates across all categories, especially on hard compositional constraints (Counting $0.90$, Position $0.83$, Attribute Binding $0.72$).
This validates our motivation for designing the tokenization around SigLIP~\citep{siglip}: by quantizing dense, semantically rich features, the resulting discrete draft preserves high-level contextual meaning while remaining fully compatible with autoregressive generation.
Consequently, both the grounded evaluator and the diffusion generator can better leverage the draft as a high-level plan, enabling more accurate diagnosis and targeted correction than VAE- or VQ-based drafts.

\begin{table*}[t]
    \centering
    \caption{
    \textbf{Comparison of Visual Draft Variants.} We compare our SigLIP-based discretization against continuous VAE latents and reconstruction-optimized VQ tokens.
    } 
   \resizebox{\linewidth}{!}{
    \begin{tabular}{c|c|cccccc}
    Draft Variants& Overall& Single Obj.& Two Obj. & Counting& Colors& Position& Attr. Binding \\
    \shline
    None (text-only) & 0.79&0.98&0.90 & 0.65&0.88&0.69& 0.61  \\
    VAE~\citep{flux} &0.72&0.98&0.86&0.60&0.82&0.54&0.52 \\
    VQ~\citep{emu3} & 0.84&0.99&0.93&0.74&0.82&0.78&0.69 \\
    \baseline{SigLIP-based Discretization} & \baseline{0.88} & \baseline{0.99} & \baseline{0.94} & \baseline{0.90} & \baseline{0.92} & \baseline{0.83} & \baseline{0.72}\\
    \end{tabular}}
    \label{tab:vq}
\end{table*}

\section{Conclusion}
We presented \method{}, leveraging the LLM as a universal reasoner for text-to-image generation that narrows the understanding--generation gap in modern unified models.
Our key insight is that with the same LLM generation may fail to avoid prompt-image inconsistencies but understanding is substantially more reliable at evaluating them.
To turn this evaluation strength into actionable control, \method{} follows a Draft-Evaluate-Diffuse pipeline: an LLM first produces the visual draft as a coarse visual plan, then performs self-critique to generate a grounded evaluation that explicitly highlights mismatches between the prompt and the drafts, and finally a diffusion model generates the final image conditioned on the joint signals (prompt, draft, evaluation) to enable targeted correction during denoising. Extensive experiments show that \method{} consistently improves semantic faithfulness and compositional constraint satisfaction under the same diffusion backbone while maintaining image quality.
We hope this work encourages a broader view of generative modeling, where the LLM serves not just as a text encoder, but as a universal reasoner that steers visual synthesis through explicit, grounded corrective signals.

\applefootnote{ \textcolor{textgray}{\sffamily Apple and the Apple logo are trademarks of Apple Inc., registered in the U.S. and other countries and regions.}}

\bibliographystyle{plainnat}
\bibliography{biblio}

\begin{thebibliography}{44}
\providecommand{\natexlab}[1]{#1}
\providecommand{\url}[1]{\texttt{#1}}
\expandafter\ifx\csname urlstyle\endcsname\relax
  \providecommand{\doi}[1]{doi: #1}\else
  \providecommand{\doi}{doi: \begingroup \urlstyle{rm}\Url}\fi

\bibitem[Bai et~al.(2023)Bai, Bai, Chu, Cui, Dang, Deng, Fan, Ge, Han, Huang, et~al.]{qwen}
Jinze Bai, Shuai Bai, Yunfei Chu, Zeyu Cui, Kai Dang, Xiaodong Deng, Yang Fan, Wenbin Ge, Yu~Han, Fei Huang, et~al.
\newblock Qwen technical report.
\newblock \emph{arXiv preprint arXiv:2309.16609}, 2023.

\bibitem[Bai et~al.(2025)Bai, Cai, Chen, Chen, Chen, Cheng, Deng, Ding, Gao, Ge, et~al.]{qwenvl}
Shuai Bai, Yuxuan Cai, Ruizhe Chen, Keqin Chen, Xionghui Chen, Zesen Cheng, Lianghao Deng, Wei Ding, Chang Gao, Chunjiang Ge, et~al.
\newblock Qwen3-vl technical report.
\newblock \emph{arXiv preprint arXiv:2511.21631}, 2025.

\bibitem[Betker et~al.(2023)Betker, Goh, Jing, Brooks, Wang, Li, Ouyang, Zhuang, Lee, Guo, et~al.]{dalle3}
James Betker, Gabriel Goh, Li~Jing, Tim Brooks, Jianfeng Wang, Linjie Li, Long Ouyang, Juntang Zhuang, Joyce Lee, Yufei Guo, et~al.
\newblock Improving image generation with better captions.
\newblock \emph{Computer Science. https://cdn. openai. com/papers/dall-e-3. pdf}, 2\penalty0 (3):\penalty0 8, 2023.

\bibitem[Chen et~al.(2025{\natexlab{a}})Chen, Xu, Pan, Hu, Qin, Goldstein, Huang, Zhou, Xie, Savarese, et~al.]{blip3o}
Jiuhai Chen, Zhiyang Xu, Xichen Pan, Yushi Hu, Can Qin, Tom Goldstein, Lifu Huang, Tianyi Zhou, Saining Xie, Silvio Savarese, et~al.
\newblock Blip3-o: A family of fully open unified multimodal models-architecture, training and dataset.
\newblock \emph{arXiv preprint arXiv:2505.09568}, 2025{\natexlab{a}}.

\bibitem[Chen et~al.(2025{\natexlab{b}})Chen, Cai, Chen, Chen, Ji, Wang, Yang, and Wang]{sharegpt4o}
Junying Chen, Zhenyang Cai, Pengcheng Chen, Shunian Chen, Ke~Ji, Xidong Wang, Yunjin Yang, and Benyou Wang.
\newblock Sharegpt-4o-image: Aligning multimodal models with gpt-4o-level image generation.
\newblock \emph{arXiv preprint arXiv:2506.18095}, 2025{\natexlab{b}}.

\bibitem[Chen et~al.(2025{\natexlab{c}})Chen, Wu, Liu, Pan, Liu, Xie, Yu, and Ruan]{chen2025janus}
Xiaokang Chen, Zhiyu Wu, Xingchao Liu, Zizheng Pan, Wen Liu, Zhenda Xie, Xingkai Yu, and Chong Ruan.
\newblock Janus-pro: Unified multimodal understanding and generation with data and model scaling.
\newblock \emph{arXiv preprint arXiv:2501.17811}, 2025{\natexlab{c}}.

\bibitem[Deng et~al.(2025)Deng, Zhu, Li, Gou, Li, Wang, Zhong, Yu, Nie, Song, et~al.]{bagel}
Chaorui Deng, Deyao Zhu, Kunchang Li, Chenhui Gou, Feng Li, Zeyu Wang, Shu Zhong, Weihao Yu, Xiaonan Nie, Ziang Song, et~al.
\newblock Emerging properties in unified multimodal pretraining.
\newblock \emph{arXiv preprint arXiv:2505.14683}, 2025.

\bibitem[Esser et~al.(2021)Esser, Rombach, and Ommer]{esser2021taming}
Patrick Esser, Robin Rombach, and Bj{\"o}rn Ommer.
\newblock Taming transformers for high-resolution image synthesis.
\newblock In \emph{CVPR}, 2021.

\bibitem[Esser et~al.(2024)Esser, Kulal, Blattmann, Entezari, M{\"u}ller, Saini, Levi, Lorenz, Sauer, Boesel, et~al.]{sd3}
Patrick Esser, Sumith Kulal, Andreas Blattmann, Rahim Entezari, Jonas M{\"u}ller, Harry Saini, Yam Levi, Dominik Lorenz, Axel Sauer, Frederic Boesel, et~al.
\newblock Scaling rectified flow transformers for high-resolution image synthesis.
\newblock In \emph{ICML}, 2024.

\bibitem[Feng et~al.(2023)Feng, Zhu, Fu, Jampani, Akula, He, Basu, Wang, and Wang]{layoutgpt}
Weixi Feng, Wanrong Zhu, Tsu-jui Fu, Varun Jampani, Arjun Akula, Xuehai He, Sugato Basu, Xin~Eric Wang, and William~Yang Wang.
\newblock Layoutgpt: Compositional visual planning and generation with large language models.
\newblock \emph{NeurIPS}, 2023.

\bibitem[Gani et~al.(2024)Gani, Bhat, Naseer, Khan, and Wonka]{gani2024llm}
Hanan Gani, Shariq~Farooq Bhat, Muzammal Naseer, Salman Khan, and Peter Wonka.
\newblock Llm blueprint: Enabling text-to-image generation with complex and detailed prompts.
\newblock In \emph{ICLR}, 2024.

\bibitem[Ghosh et~al.(2023)Ghosh, Hajishirzi, and Schmidt]{ghosh2023geneval}
Dhruba Ghosh, Hannaneh Hajishirzi, and Ludwig Schmidt.
\newblock Geneval: An object-focused framework for evaluating text-to-image alignment.
\newblock \emph{NeurIPS}, 2023.

\bibitem[Han et~al.(2025)Han, Chen, Zhao, Wang, Zhao, Yang, He, Yue, and Jiang]{tatok}
Jiaming Han, Hao Chen, Yang Zhao, Hanyu Wang, Qi~Zhao, Ziyan Yang, Hao He, Xiangyu Yue, and Lu~Jiang.
\newblock Vision as a dialect: Unifying visual understanding and generation via text-aligned representations.
\newblock \emph{arXiv preprint arXiv:2506.18898}, 2025.

\bibitem[He et~al.(2025)He, Cheng, Ma, Jia, Liu, Ma, Wu, Wu, Leng, and Yin]{he2025plangen}
Runze He, Bo~Cheng, Yuhang Ma, Qingxiang Jia, Shanyuan Liu, Ao~Ma, Xiaoyu Wu, Liebucha Wu, Dawei Leng, and Yuhui Yin.
\newblock Plangen: Towards unified layout planning and image generation in auto-regressive vision language models.
\newblock In \emph{ICCV}, 2025.

\bibitem[Ho et~al.(2020)Ho, Jain, and Abbeel]{ddpm}
Jonathan Ho, Ajay Jain, and Pieter Abbeel.
\newblock Denoising diffusion probabilistic models.
\newblock \emph{NeurIPS}, 2020.

\bibitem[Hu et~al.(2024)Hu, Wang, Fang, Fu, Cheng, and Yu]{dpg}
Xiwei Hu, Rui Wang, Yixiao Fang, Bin Fu, Pei Cheng, and Gang Yu.
\newblock Ella: Equip diffusion models with llm for enhanced semantic alignment.
\newblock \emph{arXiv preprint arXiv:2403.05135}, 2024.

\bibitem[Hurst et~al.(2024)Hurst, Lerer, Goucher, Perelman, Ramesh, Clark, Ostrow, Welihinda, Hayes, Radford, et~al.]{gpt4o}
Aaron Hurst, Adam Lerer, Adam~P Goucher, Adam Perelman, Aditya Ramesh, Aidan Clark, AJ~Ostrow, Akila Welihinda, Alan Hayes, Alec Radford, et~al.
\newblock Gpt-4o system card.
\newblock \emph{arXiv preprint arXiv:2410.21276}, 2024.

\bibitem[Karras et~al.(2022)Karras, Aittala, Aila, and Laine]{elucidating}
Tero Karras, Miika Aittala, Timo Aila, and Samuli Laine.
\newblock Elucidating the design space of diffusion-based generative models.
\newblock \emph{NeurIPS}, 2022.

\bibitem[Labs(2024)]{flux}
Black~Forest Labs.
\newblock Flux.
\newblock \url{https://github.com/black-forest-labs/flux}, 2024.

\bibitem[Li et~al.(2025)Li, Kallidromitis, Gokul, Koneru, Kato, Kozuka, and Grover]{li2025reflect}
Shufan Li, Konstantinos Kallidromitis, Akash Gokul, Arsh Koneru, Yusuke Kato, Kazuki Kozuka, and Aditya Grover.
\newblock Reflect-dit: Inference-time scaling for text-to-image diffusion transformers via in-context reflection.
\newblock In \emph{ICCV}, 2025.

\bibitem[Lian et~al.(2023)Lian, Li, Yala, and Darrell]{lian2023llm}
Long Lian, Boyi Li, Adam Yala, and Trevor Darrell.
\newblock Llm-grounded diffusion: Enhancing prompt understanding of text-to-image diffusion models with large language models.
\newblock \emph{arXiv preprint arXiv:2305.13655}, 2023.

\bibitem[Lin et~al.(2025)Lin, Li, Cheng, Niu, Ye, He, Yuan, Yu, Wang, Ge, et~al.]{uniworld}
Bin Lin, Zongjian Li, Xinhua Cheng, Yuwei Niu, Yang Ye, Xianyi He, Shenghai Yuan, Wangbo Yu, Shaodong Wang, Yunyang Ge, et~al.
\newblock Uniworld-v1: High-resolution semantic encoders for unified visual understanding and generation.
\newblock \emph{arXiv preprint arXiv:2506.03147}, 2025.

\bibitem[Loshchilov and Hutter(2017)]{adamw}
Ilya Loshchilov and Frank Hutter.
\newblock Decoupled weight decay regularization.
\newblock \emph{arXiv preprint arXiv:1711.05101}, 2017.

\bibitem[Lu et~al.(2023)Lu, Clark, Lee, Zhang, Khosla, Marten, Hoiem, and Kembhavi]{lu2023uio2}
Jiasen Lu, Christopher Clark, Sangho Lee, Zichen Zhang, Savya Khosla, Ryan Marten, Derek Hoiem, and Aniruddha Kembhavi.
\newblock Unified-io 2: Scaling autoregressive multimodal models with vision, language, audio, and action.
\newblock \emph{arXiv preprint arXiv:2312.17172}, 2023.

\bibitem[Ma et~al.(2024)Ma, Goldstein, Albergo, Boffi, Vanden-Eijnden, and Xie]{sit}
Nanye Ma, Mark Goldstein, Michael~S Albergo, Nicholas~M Boffi, Eric Vanden-Eijnden, and Saining Xie.
\newblock Sit: Exploring flow and diffusion-based generative models with scalable interpolant transformers.
\newblock In \emph{ECCV}, 2024.

\bibitem[Nichol et~al.(2021)Nichol, Dhariwal, Ramesh, Shyam, Mishkin, McGrew, Sutskever, and Chen]{glide}
Alex Nichol, Prafulla Dhariwal, Aditya Ramesh, Pranav Shyam, Pamela Mishkin, Bob McGrew, Ilya Sutskever, and Mark Chen.
\newblock Glide: Towards photorealistic image generation and editing with text-guided diffusion models.
\newblock \emph{arXiv preprint arXiv:2112.10741}, 2021.

\bibitem[OpenAI()]{chatgpt}
OpenAI.
\newblock Chatgpt.
\newblock \url{https://chatgpt.com}.

\bibitem[Peebles and Xie(2023)]{dit}
William Peebles and Saining Xie.
\newblock Scalable diffusion models with transformers.
\newblock In \emph{ICCV}, 2023.

\bibitem[Radford et~al.(2021)Radford, Kim, Hallacy, Ramesh, Goh, Agarwal, Sastry, Askell, Mishkin, Clark, et~al.]{clip}
Alec Radford, Jong~Wook Kim, Chris Hallacy, Aditya Ramesh, Gabriel Goh, Sandhini Agarwal, Girish Sastry, Amanda Askell, Pamela Mishkin, Jack Clark, et~al.
\newblock Learning transferable visual models from natural language supervision.
\newblock In \emph{ICML}, 2021.

\bibitem[Raffel et~al.(2020)Raffel, Shazeer, Roberts, Lee, Narang, Matena, Zhou, Li, and Liu]{t5}
Colin Raffel, Noam Shazeer, Adam Roberts, Katherine Lee, Sharan Narang, Michael Matena, Yanqi Zhou, Wei Li, and Peter~J Liu.
\newblock Exploring the limits of transfer learning with a unified text-to-text transformer.
\newblock \emph{JMLR}, 21\penalty0 (140):\penalty0 1--67, 2020.

\bibitem[Rombach et~al.(2022)Rombach, Blattmann, Lorenz, Esser, and Ommer]{sd}
Robin Rombach, Andreas Blattmann, Dominik Lorenz, Patrick Esser, and Bj\"orn Ommer.
\newblock High-resolution image synthesis with latent diffusion models.
\newblock In \emph{CVPR}, 2022.

\bibitem[Saharia et~al.(2022)Saharia, Chan, Saxena, Li, Whang, Denton, Ghasemipour, Gontijo~Lopes, Karagol~Ayan, Salimans, et~al.]{imagen}
Chitwan Saharia, William Chan, Saurabh Saxena, Lala Li, Jay Whang, Emily~L Denton, Kamyar Ghasemipour, Raphael Gontijo~Lopes, Burcu Karagol~Ayan, Tim Salimans, et~al.
\newblock Photorealistic text-to-image diffusion models with deep language understanding.
\newblock \emph{NeurIPS}, 2022.

\bibitem[Song et~al.(2020{\natexlab{a}})Song, Meng, and Ermon]{ddim}
Jiaming Song, Chenlin Meng, and Stefano Ermon.
\newblock Denoising diffusion implicit models.
\newblock \emph{arXiv preprint arXiv:2010.02502}, 2020{\natexlab{a}}.

\bibitem[Song et~al.(2020{\natexlab{b}})Song, Sohl-Dickstein, Kingma, Kumar, Ermon, and Poole]{score}
Yang Song, Jascha Sohl-Dickstein, Diederik~P Kingma, Abhishek Kumar, Stefano Ermon, and Ben Poole.
\newblock Score-based generative modeling through stochastic differential equations.
\newblock \emph{arXiv preprint arXiv:2011.13456}, 2020{\natexlab{b}}.

\bibitem[Tian et~al.(2025{\natexlab{a}})Tian, Gao, Gang, Lu, Gan, Yang, Wu, and Dehghan]{tian2025unigen1.5}
Rui Tian, Mingfei Gao, Haiming Gang, Jiasen Lu, Zhe Gan, Yinfei Yang, Zuxuan Wu, and Afshin Dehghan.
\newblock Unigen-1.5: Enhancing image generation and editing through reward unification in reinforcement learning.
\newblock \emph{arXiv preprint arXiv:2511.14760}, 2025{\natexlab{a}}.

\bibitem[Tian et~al.(2025{\natexlab{b}})Tian, Gao, Xu, Hu, Lu, Wu, Yang, and Dehghan]{tian2025unigen}
Rui Tian, Mingfei Gao, Mingze Xu, Jiaming Hu, Jiasen Lu, Zuxuan Wu, Yinfei Yang, and Afshin Dehghan.
\newblock Unigen: Enhanced training \& test-time strategies for unified multimodal understanding and generation.
\newblock \emph{arXiv preprint arXiv:2505.14682}, 2025{\natexlab{b}}.

\bibitem[Tschannen et~al.(2025)Tschannen, Gritsenko, Wang, Naeem, Alabdulmohsin, Parthasarathy, Evans, Beyer, Xia, Mustafa, et~al.]{siglip}
Michael Tschannen, Alexey Gritsenko, Xiao Wang, Muhammad~Ferjad Naeem, Ibrahim Alabdulmohsin, Nikhil Parthasarathy, Talfan Evans, Lucas Beyer, Ye~Xia, Basil Mustafa, et~al.
\newblock Siglip 2: Multilingual vision-language encoders with improved semantic understanding, localization, and dense features.
\newblock \emph{arXiv preprint arXiv:2502.14786}, 2025.

\bibitem[Van Den~Oord et~al.(2017)Van Den~Oord, Vinyals, et~al.]{vq}
Aaron Van Den~Oord, Oriol Vinyals, et~al.
\newblock Neural discrete representation learning.
\newblock \emph{NeurIPS}, 2017.

\bibitem[Wang et~al.(2024)Wang, Zhang, Luo, Sun, Cui, Wang, Zhang, Wang, Li, Yu, et~al.]{emu3}
Xinlong Wang, Xiaosong Zhang, Zhengxiong Luo, Quan Sun, Yufeng Cui, Jinsheng Wang, Fan Zhang, Yueze Wang, Zhen Li, Qiying Yu, et~al.
\newblock Emu3: Next-token prediction is all you need.
\newblock \emph{arXiv preprint arXiv:2409.18869}, 2024.

\bibitem[Wu et~al.(2025)Wu, Li, Zhou, Lin, Gao, Yan, Yin, Bai, Xu, Chen, et~al.]{qwenimage}
Chenfei Wu, Jiahao Li, Jingren Zhou, Junyang Lin, Kaiyuan Gao, Kun Yan, Sheng-ming Yin, Shuai Bai, Xiao Xu, Yilei Chen, et~al.
\newblock Qwen-image technical report.
\newblock \emph{arXiv preprint arXiv:2508.02324}, 2025.

\bibitem[Wu et~al.(2024)Wu, Lian, Gonzalez, Li, and Darrell]{wu2024self}
Tsung-Han Wu, Long Lian, Joseph~E Gonzalez, Boyi Li, and Trevor Darrell.
\newblock Self-correcting llm-controlled diffusion models.
\newblock In \emph{CVPR}, 2024.

\bibitem[Xie et~al.(2025)Xie, Chen, Chen, Cai, Tang, Lin, Zhang, Li, Zhu, Lu, et~al.]{sana}
Enze Xie, Junsong Chen, Junyu Chen, Han Cai, Haotian Tang, Yujun Lin, Zhekai Zhang, Muyang Li, Ligeng Zhu, Yao Lu, et~al.
\newblock Sana: Efficient high-resolution text-to-image synthesis with linear diffusion transformers.
\newblock In \emph{ICLR}, 2025.

\bibitem[Yang et~al.(2024)Yang, Yu, Meng, Xu, Ermon, and Cui]{yang2024mastering}
Ling Yang, Zhaochen Yu, Chenlin Meng, Minkai Xu, Stefano Ermon, and Bin Cui.
\newblock Mastering text-to-image diffusion: Recaptioning, planning, and generating with multimodal llms.
\newblock In \emph{ICML}, 2024.

\bibitem[Yu et~al.(2022)Yu, Xu, Koh, Luong, Baid, Wang, Vasudevan, Ku, Yang, Ayan, et~al.]{parti}
Jiahui Yu, Yuanzhong Xu, Jing~Yu Koh, Thang Luong, Gunjan Baid, Zirui Wang, Vijay Vasudevan, Alexander Ku, Yinfei Yang, Burcu~Karagol Ayan, et~al.
\newblock Scaling autoregressive models for content-rich text-to-image generation.
\newblock \emph{arXiv preprint arXiv:2206.10789}, 2022.

\end{thebibliography}

\end{document}